\documentclass[journal]{IEEEtran}
\IEEEoverridecommandlockouts
\usepackage{cite}
\usepackage{amsmath,amssymb,amsfonts}
\usepackage{algorithmic}
\usepackage{algorithm}
\usepackage{graphicx}
\usepackage{textcomp}
\usepackage{xcolor}
\usepackage{hhline}
\usepackage{svg}
\usepackage{booktabs}
\usepackage{multirow}
\usepackage{comment}

\def\BibTeX{{\rm B\kern-.05em{\sc i\kern-.025em b}\kern-.08em
    T\kern-.1667em\lower.7ex\hbox{E}\kern-.125emX}}
\begin{document}

\title{Flow-Based Visual Stream Compression for\\Event Cameras

\thanks{Daniel C. Stumpp is with the NSF-SHREC Center at the University of Pittsburgh, Pittsburgh, PA (e-mail: daniel.stumpp@pitt.edu). Himanshu Akolkar is with the University of Pittsburgh, Pittsburgh, PA (e-mail: akolkar@pitt.edu). Ryad Benosman is with the University of Pittsburgh, Pittsburgh, PA (benjry.benos@gmail.com). Alan D. George is with the NSF-SHREC Center at the University of Pittsburgh, Pittsburgh, PA (email: alan.george@pitt.edu).\\
This research was supported by the NSF Center for Space, High-performance, and Resilient Computing (SHREC) industry and agency members and by the IUCRC Program of the National Science Foundation under Grant No. CNS-1738783.\\This work has been submitted to the IEEE for possible publication. Copyright may be transferred without notice, after which this version may no longer be accessible}}

\author{\IEEEauthorblockN{Daniel C. Stumpp,}
\and
\IEEEauthorblockN{Himanshu Akolkar,}
\and
\IEEEauthorblockN{Alan D. George,}
\and
\IEEEauthorblockN{Ryad Benosman}
}

\maketitle

\begin{abstract}
As the use of neuromorphic, event-based vision sensors expands, the need for compression of their output streams has increased. While their operational principle ensures event streams are spatially sparse, the high temporal resolution of the sensors can result in high data rates from the sensor depending on scene dynamics. For systems operating in communication-bandwidth-constrained and power-constrained environments, it is essential to compress these streams before transmitting them to a remote receiver. Therefore, we introduce a flow-based method for the real-time asynchronous compression of event streams as they are generated. This method leverages real-time optical flow estimates to predict future events without needing to transmit them, therefore, drastically reducing the amount of data transmitted. The flow-based compression introduced is evaluated using a variety of methods including spatiotemporal distance between event streams. The introduced method itself is shown to achieve an average compression ratio of 2.81 on a variety of event-camera datasets with the evaluation configuration used. That compression is achieved with a median temporal error of 0.48~$ms$ and an average spatiotemporal event-stream distance of 3.07. When combined with LZMA compression for non-real-time applications, our method can achieve state-of-the-art average compression ratios ranging from 10.45 to 17.24. Additionally, we demonstrate that the proposed prediction algorithm is capable of performing real-time, low-latency event prediction.
\end{abstract}

\begin{IEEEkeywords}
Event-Based, Compression, Optical Flow, Neuromorphic, Low-Power
\end{IEEEkeywords}

\section{Introduction}
\label{sec:introduction}
Event-based vision sensors perform inherent compression of visual scene information by only transmitting events when the illuminance at a given pixel changes \cite{Brandli2014ASensor}. While events are spatially sparse due to this inherent compression, the high temporal resolution of event-based vision sensors results in high event rates in active areas of the scene. These high event rates can become prohibitive, especially for embedded platforms with significant power constraints. To overcome high event rates, compression of these event streams has gained increased interest. 

Like traditional image or video compression, event-stream compression can be classified as lossless or lossy depending on whether any information is lost between the original event stream and the decompressed event stream. In this research, we present a lossy event-stream compression technique that uses asynchronous event-based optical flow to reduce data transmission rates by enabling accurate stream reconstruction through flow-based prediction at the receiver. The flow-based compression method introduced here has two primary characteristics that make it advantageous compared to other methods in literature: (1) it can be asynchronously applied to events as they are generated and does not inherently require time aggregation of events and (2) it is a stream-to-stream method for compression meaning that the compressed data itself is an event stream to which further compression may be applied if desired and it can be used as input to already deployed event-based algorithms. In addition to these two characteristics, the proposed flow-based compression is also designed to have very low compression overhead to enable deployment on resource- and power-constrained edge devices. There are a variety of related techniques that have been used for the compression of event streams, however, the stream-to-stream nature of the proposed method means that these existing techniques and the new method introduced are not mutually exclusive; both may be applied and the benefits of each gained. 

The following sections will outline the introduced flow-based compression (FBC) technique and demonstrate its capabilities for the compression of event streams. Sections~\ref{sec:related-research} and \ref{sec:background} explore existing methods for event stream compression along with the principles behind event-based vision sensors and optical flow. Methods and results are explored in Sections~\ref{sec:methods} and \ref{sec:experiments-and-results} followed by comparisons to existing research and conclusions about the feasibility and effectiveness of flow-based compression.

\section{Related Research}
\label{sec:related-research}

Research regarding the compression of event streams has increased in recent years. We broadly classify existing methods into two groups for discussion: lossless and lossy. The existing methods within these categories will be discussed. Types of compression considered include both event-stream specific methods along with general purpose methods that have been applied to event-streams.

\subsection{Lossless Compression Methods} \label{sec:RW:lossless}
Lossless compression methods are those for which the decompressed or reconstructed event stream is identical to the original stream before compression. A lossless compression method using spike coding was introduced in \cite{bi2018spike} and expanded on in \cite{dong2018spike}. This method divides the event stream into multiple spatiotemporal event cubes, which are then individually encoded by capturing the spatial and temporal redundancies in each of the local event stream cubes. The cubes are encoded using both adaptive octree-based cube partitioning and intracube prediction. While the spike compression method was specific to the compression of event streams, some other explorations of event-stream compression focused on utilizing existing general-purpose methods of compression.

An exploration of general-purpose compression techniques for event-stream compression was presented in \cite{khan2020lossless}. They explored methods such as entropy coding, dictionary-based compression, fast integer compression, and internet-of-things-specific compression (e.g., Sprintz \cite{blalock2018sprintz}). The use of these methods was compared to the spike coding method in \cite{dong2018spike}. They find that the dictionary-based Lempel-Ziv-Markov chain algorithm (LZMA) provides the highest compression ratio when the sensor is static and that the spike-coding method provides the highest compression ratio when the sensor is in motion. However, they note the high complexity of both of these methods, with spike coding being particularly time-consuming to perform. Because of these high complexities, they recommend the dictionary-based Brotli compression algorithm \cite{alakuijala2018brotli}, as it has the best trade-offs between compression ratio and complexity for both static and dynamic event stream datasets.

In \cite{schiopu2022lossless}, a compression method is introduced using bitmap encoding of aggregated event frame representations. The large temporal windows used for aggregating event frames and encoding make this technique primarily applicable for the storage of event streams or non-real-time processing scenarios. The same authors \cite{schiopu2022low, schiopu2022low2,schiopu2023entropy} explored compression techniques based on a triple thresholding partition (TTP) algorithm that is used to divide the input stream into different coding ranges. In \cite{schiopu2023entropy}, an improved LLC-ARES entropy-based coding is used and shown to have improved runtime compared to LZMA. Because this method aggregates events and works on rearranged event sequences, it too is more suitable for applications that do not need to operate on events asynchronously in real time. The research in \cite{schiopu2022low} and \cite{schiopu2022low2} focused on the use of the TTP algorithm for applications on low-power hardware. Although these methods are able to provide efficient methods for the storage of event data even on resource-constrained devices, they are not directly applicable to the goal of the flow-based compression introduced in this paper: to perform compression asynchronously in real-time as events are generated.

While most of the previously discussed methods focused on accumulating and compressing event-frame representations, there has also been work exploring the direct compression of point-cloud representations. In \cite{martini2022lossless}, the authors first generated 3-D point clouds for each polarity of events. These point clouds are aggregated over time periods on the order of one second. The geometry-based point-cloud compression (G-PCC) codec from MPEG is then applied to both of the point clouds in parallel to generate the compressed output. The research in \cite{huang2023event} also utilizes a point-cloud representation for event-stream compression. In this case, the polarity is directly encoded with the data in the point cloud. Multiple point-cloud compressors are explored, but they ultimately settle on G-PCC as well. Different time-splitting methods are evaluated including using fixed time intervals and a fixed number of events. It is found that using a fixed number of events results in better performance because it ensures more spatial redundancy in the data \cite{huang2023event}. Overall, like the other lossless methods discussed, the point-cloud methods cannot fulfill the real-time asynchronous processing requirement due to substantial time aggregation of events.

\subsection{Lossy Compression Methods} \label{sec:RW:lossy}
Lossy compression methods are those where the reconstructed event stream is not guaranteed to be exactly identical to the original event stream before compression. One such method for lossy compression is known as time-aggregation-based lossless video encoding for neuromorphic vision sensors (TALVEN) \cite{khan2020time}. Although the name implies that this method is lossless, it relies on lossy time aggregation of events such that the exact time of a given event cannot be recovered with the original temporal resolution of the sensor. Therefore, we consider this to be a lossy method that will incur temporal error upon event-stream reconstruction. TALVEN relies on time aggregation of events followed by frame coding. TALVEN utilizes existing video-encoding methods on aggregated superframes over an aggregation period of up to 50 $ms$. The point-cloud approach in \cite{martini2022lossless} expands on TALVEN and through the use of point clouds prevents the loss of temporal information.

Another method for lossy event-stream compression is the use of quadtrees and Poisson disk sampling as introduced in \cite{banerjee2021lossy}. This method achieved a compression ratio of more than 6$\times$ those presented in prior works. Although this work achieves a high compression ratio, it is limited by its dependence on the use of intensity images generated by an event-based vision sensor such as the DAVIS or RGB-DAVIS \cite{banerjee2021lossy}. This limitation is a significant drawback as it cannot be used with purely asynchronous temporal-contrast sensors, but instead has to rely on additional intensity data to select the low-priority regions where events can be removed without impacting the quality of event reconstruction. 

A more recent lossy compression method presented in \cite{khaidem2022novel} uses machine learning to compress an event stream. A deep belief network (DBN) is used to reduce the input stream into a latent representation that is then encoded using an entropy-based encoding method. This method uses time-aggregated superframes similar to TALVEN. Both the time aggregation and the DBN ending are lossless. The time aggregation specifically results in a significant loss of temporal resolution. However, a high peak signal-to-noise ratio of decompressed frames is demonstrated \cite{khaidem2022novel}. This machine-learning-based approach demonstrates high compression ratios of up to 327 on the datasets used. These compression ratios, however, are dependent on the aggregation period for the frames used. When short time aggregation periods ($<5 ms$) are used the compression ratio drops to as low as 0.84. The inference time required to process data through the DBN and then the entropy encoding could also be prohibitive to the compression of event streams on the edge. Latency is dependent on sensor size, time aggregation and other parameters discussed in \cite{khaidem2022novel}.

Recently, a lossy point-cloud compression approach has been explored in \cite{huang2023evaluation}. The authors expand their lossless compression method in \cite{huang2023event} by updating the parameters of the G-PCC algorithm to make it lossy. They then evaluate the output event streams on various event-based processing tasks such as classification, optical flow estimation, and depth estimation. Overall, from their results, it is concluded that the variations in the event streams caused by lossy compression are warranted due to the improved compression ratios achieved. The research in \cite{huang2023evaluation} does not use quantitative comparisons between the original event streams and the reconstructed event streams. Only task-based evaluation is performed. While this does provide good insight into the applicability of the lossy method it fails to provide an understanding of the magnitude of deviations between the original and reconstructed event streams. In this research, we will explore quantitative metrics for the evaluation of event-stream dissimilarities after lossy reconstruction.

\section{Background} \label{sec:background}
The following sections provide background related to the flow-based compression method introduced in this research. First, the operation of event-based vision sensors is discussed. Then a brief survey of methods used for the computation of event-based optical flow is presented.

\subsection{Event-Based Vision Sensors}
\label{sec:BG:event-based-sensors}

Event-based vision sensors are neuromorphic sensors inspired by the biological operation of the eye. Instead of sampling the intensity of all pixels at a synchronized frame rate, event-based vision sensors record asynchronous events triggered by changes in scene intensity. The events triggered are either ``ON'' or ``OFF'' corresponding to increasing or decreasing temporal log intensity over time, respectively \cite{Gallego2020Event-basedSurvey}. Events are output in the address-event representation (AER) format. The AER format packs the event location, time, and polarity into an 8-byte ($x$, $y$, $t$, $p$) packet \cite{Boahen2000Point-to-pointEvents}.

The operating paradigm of event-based vision sensors brings multiple advantages when compared to traditional cameras. Most notably, event-based vision sensors provide high dynamic range due to the autonomous nature of each pixel as well as microsecond temporal resolution due to the asynchronous triggering of events \cite{Gallego2020Event-basedSurvey}. The high dynamic range enables the sensor to detect and track objects in lighting environments where traditional cameras would be saturated or underexposed. The high temporal resolution of the sensors enables the accurate and fast computation of optical flow---a capability that is leveraged in the flow-based compression introduced in this research. In addition to the high dynamic range and temporal resolution, these sensors also produce sparse data as they perform inherent spatial compression \cite{Brandli2014ASensor}. This compression capability is advantageous for the reduction of sensor data rates, however, the data rates can still become prohibitively large due to the high temporal resolution of events being triggered.

\subsection{Event-Based Optical Flow}
\label{sec:BG:event-based-flow}

Optical flow is the estimation of motion, traditionally from sequential frames of imagery. Event-based optical flow computes estimated motion based on the high temporal resolution, asynchronous events generated by event-based vision sensors. Event-based optical flow can be computed sparsely for each event or densely across the whole frame based on events. Additionally, optical flow can be computed as the local flow or the true flow. Local flow, such as the regularized plane fitting approach in \cite{Benosman2014Event-basedFlow}, produces a flow vector that is always normal to the moving edge, regardless of the true direction of motion. This phenomenon is known as the aperture problem and limits the effectiveness of many event-based optical flow estimation techniques by preventing them from accurately estimating the true magnitude and direction of motion (true flow). The flow-based compression introduced in this research is reliant on the ability to asynchronously compute real-time optical flow that is accurate in both magnitude and direction. 

There have been a variety of optical-flow algorithms introduced to address the aperture problem and compute true flow. A block-matching method for computing flow is presented and implemented on an FPGA in \cite{Liu2017Block-matchingImplementation,Liu2019AdaptiveSensors}. However, this method does not asynchronously compute the flow for each event, instead accumulating events into frames before performing the block matching. An aperture robust method for computing flow that relies on multi-scale pooling of local-flow estimates referred to as ARMS was presented in \cite{Akolkar2020Real-timeFlow}. In previous research, we presented the faster ARMS (fARMS) software-optimized version of ARMS as well as the hardware ARMS (hARMS) architecture for high-speed event-based optical flow on an FPGA. The hARMS architecture is able to achieve up to 25\% reduction in average endpoint error while computing the event-based optical flow asynchronously at up to 1.21 Mevent/s \cite{stumpp2021harms}. There have also been machine-learning approaches to generating event-based true flow. Two notable methods are EV-FlowNet \cite{Zhu2018EV-FlowNet:Cameras} and E-RAFT \cite{gehrig2021raft}. EV-FlowNet uses a self-supervised neural network to achieve good accuracy but is dependent on the use of accumulated frames to estimate the flow. It is also dependent on the use of DAVIS grayscale images for training, making retraining and adaption to new scene dynamics more difficult. E-RAFT introduces an architecture that considers recurrence and temporal priors based on the RAFT architecture presented in \cite{teed2020raft}. Using this architecture E-RAFT is able to achieve state-of-the-art flow estimation accuracy and also compute dense optical flow throughout the whole scene, even where events aren't active. Like EV-FlowNet, E-RAFT uses temporal aggregation of events before computing flow. In the case of E-RAFT events are collected into event packets that are stored in tensors and used as input to the model.

\section{Methods}
\label{sec:methods}

\begin{figure}[t]
    \centering
    \includegraphics[width=\linewidth]{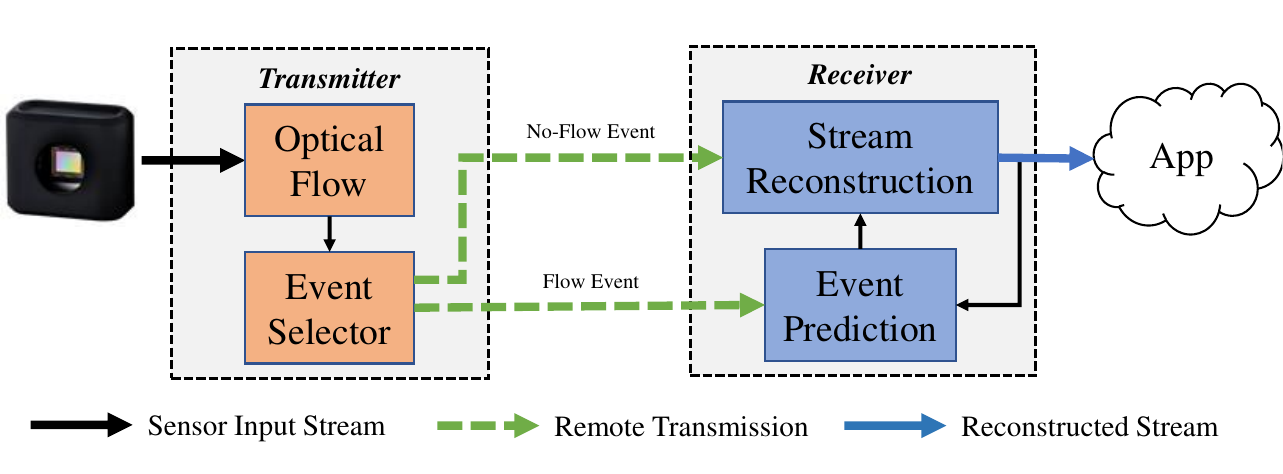}
    \caption{Conceptual diagram for flow-based compression system.}
    \label{fig:FBC-conceptual}
\end{figure}

The structure of the Flow-Based Compression (FBC) system introduced here is composed of two primary components---the \textit{transmitter} and the \textit{receiver}---as shown in Fig.~\ref{fig:FBC-conceptual}. The transmitter is located `on the edge' and connected to an event-based vision sensor, while the receiver may be another edge device or, more likely, a more powerful computing platform. We, therefore, aim to minimize the transmitter-side computation to allow deployment on devices with minimal computing resources. As seen in Fig.~\ref{fig:FBC-conceptual}, events are passed from the sensor to the transmitter where optical flow is calculated using a method such as hARMS, and the events to send are selected. A confidence value is computed for each flow event. If the confidence of flow computation is high i.e. the calculated flow is consistent and can be used reliably to perform prediction, then these events are sent alongside the calculated flow. Events that do not have confident flow computation are sent as is without the flow information. The selected events and their flow are used by the receiver to perform event prediction and stream reconstruction (decompression). The receiver uses the transmitted flow events to predict the spatiotemporal location of future events. If future events can be accurately predicted, the event stream can be compressed by simply not transmitting future events for a period of time after a flow event is received. The details of this prediction algorithm are discussed in Section~\ref{sec:event-prediction} after the FBC system timing is introduced. Finally, after prediction the reconstructed event stream is asynchronously streamed into any application designed for processing event streams. This system requires less data to be transmitted via the remote connection between the transmitter and the receiver, thus saving power and bandwidth. In the following sections, the methodologies for system timing, event prediction, and stream reconstruction are presented. System complexity is also discussed and evaluation metrics used in this research are introduced.

\subsection{System Timing}
\label{sec:system-timing}

The proposed compression algorithm has a simple timing structure consisting of two states: \textit{sending} and \textit{predicting}. The sending phase is the duration when the flow is calculated by the transmitter and sent to the receiver. If we assume that the predictions at the receiver are performed correctly then the transmitter does not need to send any more events during the subsequent predicting phase. This simple two-state design enables low transmitter-side complexity and minimizes communication overhead for timing and synchronization. Further motivation for having two system states as well as details of each state are as follows.

The calculation of flow from an event represents an estimation of the instantaneous direction and magnitude of motion for that event on the sensor's focal plane. In the case of an ideal sensor observing motion with constant magnitude and direction that flow estimate can be used to predict the location of the event indefinitely into the future. However, due to sensor non-idealities, errors in the flow estimate, and variable scene dynamics, the error between the actual event location and the predicted event location based on the flow estimate will begin to increase as time progresses. There are at least two ways to address this problem. The first is for the transmitter to track the flow for each event that occurs as it moves through the focal plane and update the receiver with new flow information as required to maintain the desired level of prediction accuracy. The second approach is to use a periodic update schedule that transmits event and flow information on a pre-determined (but adjustable) time interval for all events in the scene. 

The first approach allows for event-by-event control over the compression quality and the transmission of flow in a totally asynchronous way. However, it is faced with two significant problems. First, the challenge of matching a current event to the appropriate corresponding event in the past is difficult. In practice, this is often impossible due to sensor properties, rapid changes in the scene, and the merging and splitting of events. Even if an appropriate matching algorithm were devised, this would directly lead to the second problem: unacceptable computational complexity. The goal of this system, as previously stated, is to keep the transmitter processing within the scope of what could be reasonably handled by embedded systems that are likely to deploy with an event-based sensor. This method for updating flow, however, introduces significant complexity to the transmitter, requiring it to compute flow, perform predictions, and match events. These limitations led to the selection of the second option; a periodic update scheme that leverages two system states.

The \textit{sending} state is a period of time ($ST$) in which all events and their flow are being transmitted from the transmitter to the receiver. The \textit{predicting} state is the period of time ($PT$) during which most of the new events added to the reconstructed stream are generated using flow-based predictions. In the \textit{predicting} state, events for which a flow cannot be calculated are transmitted to the receiver under the assumption that predictions will only be made for the events that would have had a flow estimate on the receiver side. After $PT$ of predicting events the next \textit{sending} period begins, the event buffers on the receiver are cleared and new events and flow estimates are received. Of these two parameters, only $PT$ must be explicitly configured; $ST$ is adaptively computed as discussed in Section~\ref{sec:param-config}.

\subsection{Event Prediction}
\label{sec:event-prediction}

When the system is in the $predicting$ state the transmitter is only sending events for which it cannot accurately estimate an optical flow. At the same time, the receiver is collecting events and performing predictions for the events that were transmitted during the $sending$ state. This prediction aims to reconstruct the event stream using the optical flow estimates received during the previous $sending$ state. This task presents some challenges that must be considered in order to effectively reconstruct the event stream.

Each flow event transmitted during the $sending$ state is used to perform flow-based predictions. The flow-based predictions are performed with three primary underlying assumptions: (1) that the estimated optical flow accurately reflects the true direction and magnitude of motion on the focal plane, (2) that the magnitude and direction of motion do not change substantially throughout the duration of the $predicting$ state, and (3) that the cause of the event may be approximated by a point source located at the center of the pixel at the time the event is triggered. Based on these assumptions, we formulate an analytical solution for computationally efficient and accurate prediction of future events. The solution is formulated in three dimensions: $x$ and $y$ in pixel space, and time $t$. The subpixel trajectory is the predicted path of the event based on the optical-flow estimate. Pixel locations are represented as lines in three-dimensional space and change only in time. The objective then, is to find the time $t_{min}$ for each potential event pixel that minimizes the distance between the pixel location line and the subpixel trajectory line. If this minimum distance is less than a defined pixel slack ($\xi$), then an event is predicted to occur at that pixel at the time $t_{min}$. In practice, we minimize the distance squared to improve computational efficiency. The distance squared between a pixel line $(x_p, y_p, t)$ where $x_p$ and $y_p$ are constant and subpixel trajectory is shown by Eq.~\ref{eq:dist-2}. Note that there is no time component of distance because we are minimizing distance subject to the constraint that $t_1 = t_2$.

\begin{equation}
    \label{eq:dist-2}
    d^2 = (v_xt - x_p)^2 + (v_yt - y_p)^2
\end{equation}

Taking the derivative of Eq.~\ref{eq:dist-2} with respect to time yields Eq.~\ref{eq:dd2}. Minimizing this equation in Eq.~\ref{eq:tmin1} and Eq.~\ref{eq:tmin2} yields the solution for $t_{min}$. The value of $t_{min}$ is a function only of the event velocity and the candidate event pixel coordinates being considered, all of which are constants.

\begin{equation}
    \label{eq:dd2}
    \frac{d(d^2)}{dt} = 2v_x(v_xt - x_p) + 2v_y(v_yt - y_p)
\end{equation}

\begin{equation}
    \label{eq:tmin1}
    2v_x(v_xt_{min} - x_p) + 2v_y(v_yt_{min} - y_p) = 0
\end{equation}

\begin{equation}
    \label{eq:tmin2}
    t_{min} = \frac{2v_yy_p + 2v_xx_p}{2v_x^2 + 2v_y^2}
\end{equation}

\begin{figure}
    \centering
    \includegraphics[width=\linewidth]{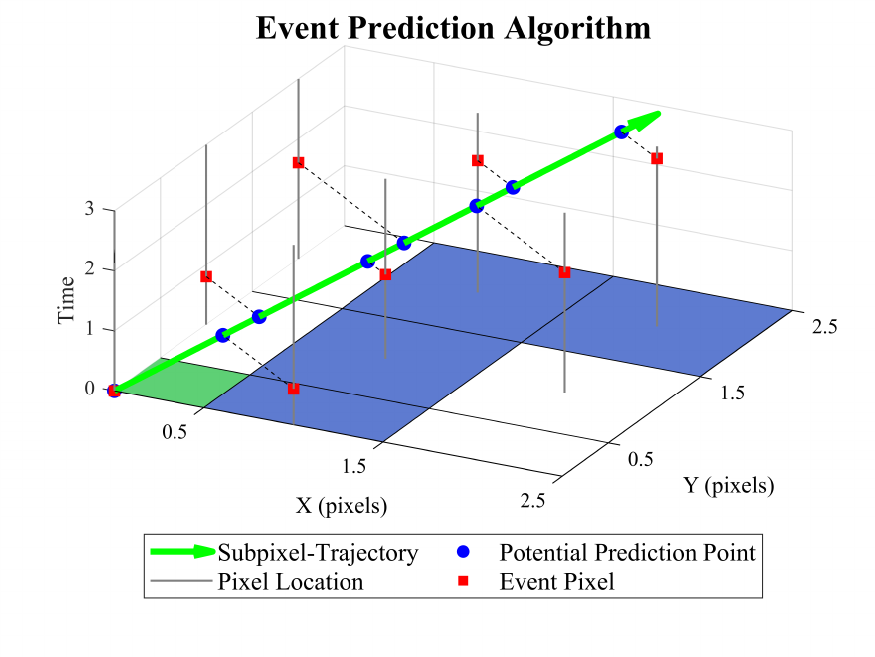}
    \caption{Visualization of the event prediction algorithm. Shaded pixel locations represent the initial (green) and predicted (blue) events generated by the algorithm.}
    \label{fig:3d-pred-alg}
\end{figure}

Once $t_{min}$ is directly computed, it is used to solve for $d^2$ using Eq.~\ref{eq:dist-2}. This distance is then compared to $\xi$ to determine if an event should be triggered. Note that throughout the calculations it is assumed that the event prediction begins at the origin. The three-dimensional formulation of the prediction problem is visualized in Fig.~\ref{fig:3d-pred-alg} where the blue shaded pixels represent the predicted event locations.

\begin{algorithm}
    \small
    \caption{Modified Bresenham algorithm for electing possible pixels.}
    \label{alg:modified-bresenham}
    \begin{algorithmic}[1]
        \STATE \textbf{function} \textsc{modified\_bresenham}$(x0, y0, x1, y1)$
            \STATE $xDist = $\textsc{abs}$(x1 - x0)$
            \STATE $yDist = -$\textsc{abs}$(y1 - y0)$
            \STATE $xStep = (x0 < x1~?~1~:~-1)$
            \STATE $xStep = (y0 < y1~?~1~:~-1)$
            \STATE $error = xDist + yDist$

            \STATE $coords \longleftarrow$ empty point vector
            \WHILE{$x0 \neq x1$ \textbf{or} $y0 \neq y1$}
                \IF{$2\cdot error - yDist > xDist - 2\cdot error$}
                    \STATE $error += yDist$
                    \STATE $x0 += xStep$
                    \STATE $coords.append([x0, y0])$
                    \STATE $coords.append([x0-xStep, y0+yStep])$
                \ELSE
                    \STATE $error += xDist$
                    \STATE $y0 += yStep$
                    \STATE $coords.append([x0, y0])$
                    \STATE $coords.append([x0+xStep, y0-yStep])$
                \ENDIF
            \ENDWHILE
            \RETURN $coords$
        \STATE \textbf{end function}
    \end{algorithmic}
\end{algorithm}
\begin{algorithm}
    \small
    \caption{Event prediction algorithm.}
    \label{alg:predict}
    \begin{algorithmic}[1]
        \STATE \textbf{function} \textsc{predict\_event}$(evt, send\_end)$
            \STATE $offX = evt.x$
            \STATE $offY = evt.y$
            \STATE $v_x = evt.vx$
            \STATE $v_y = evt.vy$

            \STATE $stop\_time = send\_end + PT$
            \STATE $duration = stop\_time - evt.t$
            \STATE $endX = ceil(v_x\cdot duration)$
            \STATE $endY = ceil(v_y\cdot duration)$

            \STATE $coords = $\textsc{modified\_bresenham}$(0, 0, endX, endY)$
            \STATE $preds \longleftarrow$ empty event vector
            \FORALL{$[x_p, y_p]$ \textbf{in} $coords$}
                \STATE $t_{min} = (2\cdot v_y\cdot y_p + 2\cdot vx \cdot xp) / (2\cdot v_x^2 + 2\cdot v_x^2)$
                \STATE $d2 = (v_x\cdot t_{min} - x_p)^2 + (v_y\cdot t_{min} - y_p)^2$
                \IF{$d2 < \xi^2$ \textbf{and} $evt.t + t_{min} > send\_end$}
                    \STATE $preds.append([offX + x_p, offY + y_p, evt.t + t_{min}, evt.p])$
                \ENDIF
            \ENDFOR
            \RETURN $preds$
        \STATE \textbf{end function}
    \end{algorithmic}
\end{algorithm}

The receiver requires an efficient method of determining the potential event pixel locations to be used in the prediction algorithm. To efficiently find these potential pixel coordinates we use a modified version of Bresenham's line-drawing algorithm \cite{bresenham1998algorithm} as shown in Algorithm~\ref{alg:modified-bresenham}. This thickens the line generated by the standard Bresenham algorithm to ensure that all pixels the trajectory passes over will be included as potential coordinates. This algorithm can be efficiently computed as it iterates only through the relevant points along the line and does not require floating-point math to determine the output vector of potential candidate event pixels. 

For each event with flow transmitted during the $sending$ state, the full event prediction is performed with Algorithm~\ref{alg:predict}. This algorithm uses the modified Bresenham algorithm along with Eq.~\ref{eq:tmin2} to compute the predicted events. The $send\_end$ input is the ending time of the $sending$ state in which the flow event $evt$ was sent. Note that it is possible that the same source could generate multiple events at the same pixel. This behavior is not modeled in the prediction algorithm, however, in that case, there would be multiple sent flow events in close temporal proximity and independent predictions of those events will propagate that behavior through the prediction time window.

Upon completion of event prediction for each flow event sent, the final task performed by the receiver is to sort all of the predicted events generated from the various sent flow events. Depending on the real-time constraints of the system, this sort and send of predicted events can be performed periodically on fine-grained intervals as predictions are generated. It is during this sorting step that predicted events are combined with any new events that were transmitted during the $predicting$ state due to the lack of a valid flow estimate. Sorting is performed using the IntroSort algorithm \cite{musser1997introspective}.

\subsection{Parameter Configuration}
\label{sec:param-config}

With the introduced flow-based compression algorithm in mind, we now discuss the parameter-configuration techniques used. The $ST$ of the system must be appropriately configured to ensure the number of output events matches closely the number of input events to achieve the best reconstruction accuracy. A send time that is too long will result in too many events being predicted, while a smaller than optimal $ST$ will have the opposite effect. Therefore, as previously mentioned, the $ST$ is adaptively computed at runtime. It is assumed that $ST$ is optimal for a given event when it is equivalent to the time it takes that event to travel one pixel. This prevents aliasing or duplicate event predictions. Therefore, we adaptively compute the send time for each $sending$ state according to Eq.~\ref{eq:adaptive-ST} where $C$ is the number of flow events used for calibration.

\begin{equation}
    \label{eq:adaptive-ST}
    ST = \frac{1}{\frac{1}{C}\sum_{i=0}^{C} \sqrt{v_{xi}^2 + v_{yi}^2}}
\end{equation}

In this research, we use $C = 500$ flow events to perform the calibration of $ST$ for each time in the $sending$ state. After $C$ flow events have been sampled in the $sending$ state the new $ST$ is computed and the end time of the current $sending$ state is updated accordingly. This method introduces negligible overhead to the system. This approach will see its best results for scenarios where most objects are moving at similar speeds---direction does not matter. The effectiveness of this method will be discussed in Section~\ref{sec:experiments-and-results}.

With $ST$ adaptively configured, the only remaining timing parameter to consider is the $PT$. In this research, this value is configured as a fixed time for the sake of characterizing the proposed method. However, $PT$ could also be configured adaptively based on the mean flow magnitudes, where a higher flow magnitude would decrease the adaptive $PT$ under the assumption that objects moving faster are more likely to change direction or speed than slowly moving objects. This assumption is not evaluated in this research. 

\subsection{System Complexity}
Having introduced the operational model for the flow-based compression system we now discuss the system's theoretical computational complexity. First, the computational complexity of the transmitter subsystem is discussed in the context of embedded deployment. Second, the receiver-subsystem complexity is discussed with an analysis of the flow-based prediction algorithm's complexity.

\subsubsection{Transmitter}
The transmitter-subsystem complexity is almost entirely comprised of the computation of optical flow. It was demonstrated in \cite{stumpp2021harms} that real-time performance can be achieved for event-based optical flow on embedded platforms with the FPGA accelerated architecture. EDFlow introduced in \cite{liu2022edflow} also enables real-time computation of event-based flow on FPGA, however, it does not operate purely asynchronously. Beyond the complexity of these optical-flow calculations, which is explored in detail in \cite{stumpp2021harms} and \cite{liu2022edflow}, the FBC algorithm introduces very little additional complexity on the transmitter side. The only additional workload is the operation of the system timing state machine and the event selector to determine what events are sent. The event selector is simple: only sending events during the $sending$ state or when an accurate flow estimate cannot be obtained. Therefore, the additional complexity is negligible and FBC can be deployed on resource-constrained edge devices that are capable of supporting real-time event-based flow.

\subsubsection{Receiver}
The receiver subsystem contains most of the complexity introduced by the FBC algorithm. Algorithm~\ref{alg:predict} needs to be performed for each event with flow transmitted during the $sending$ state. Therefore, the time complexity of the overall prediction scales linearly with the number of events prediction must be performed for ($N_{evt}$). The time complexity also scales with the number of potential coordinates produced by Algorithm~\ref{alg:modified-bresenham}, denoted as $N_{mb}$. The final contribution to the receiver-subsystem complexity is the sorting of all predicted events output into temporal order which has a complexity $O(N_{pred}\log(N_{pred}))$ where $N_{pred}$ is the number of predicted events. Therefore, the overall complexity can be represented as $O(N_{evt}N_{mb}N_{pred}\log(N_{pred}))$. In practice the value of $N_{mb}$ is at least an order of magnitude lower than all other factors. We assume that $N_{evt}$ and $N_{pred}$ are of the same order of magnitude as is expected except for rare cases. Therefore, the overall time complexity of the receiver subsystem is on the order $O(n^2\log(n))$ when implemented serially.

Although the absolute value of $n$ is relatively low, it is possible that the time complexity of the receiver subsystem will become prohibitive, especially for resource-constrained devices. In these cases, it is desirable to leverage algorithm parallelism. The nature of the prediction algorithm developed presents a significant opportunity for parallelism and hardware acceleration. Because each event's predictions are computed independently from other events, that process can be performed in parallel for all events. The sorting remains the primary contributor to system complexity, however, this can be improved by outputting predicted events at a small interval throughout the prediction and performing many small-batch sorts in parallel with ongoing predictions.

\subsection{Multi-Method Cascaded Compression}

Unlike existing compression methods for event streams, the FBC method introduced is unique in that it is a stream-to-stream compression approach. This method is advantageous to the primary application of interest: asynchronous, event-by-event compression for real-time applications. However, in scenarios where additional latency is acceptable, the stream-to-stream nature of FBC enables the possibility of further compression through the use of an additional event-stream-compression method. To demonstrate this use case we develop and evaluate a cascaded compression architecture where FBC is used for initial compression and the resulting stream is then further compressed using an existing method of compression. For this demonstration LZMA is used due to the high compression ratio it achieved in \cite{khan2020lossless}. This characteristic is a unique advantage of the FBC method and enables compression gains to be significantly compounded. Results and discussion of the effectiveness of cascaded compression are provided in Section~\ref{sec:combination-method}.

\subsection{Evaluation Metrics}
\label{sec:evaluation-metrics}

To evaluate the performance of the FBC method introduced we consider three key metric areas: compression, event-stream distance, and temporal error. These metrics are discussed in the following sections. The results reported in Section~\ref{sec:experiments-and-results} use the metrics outlined here.

\subsubsection{Compression}
\label{sec:metric-compression}

We use two metrics for the evaluation of the compression performance. First, we consider the event reduction (ER). This metric is defined as the percentage of events in the original stream that were never sent from the transmitter to the receiver. The equation for ER is formalized in \eqref{eq:ER} where $N_{s}$ and $N_{tx}$ are the number of events in the input stream and the number of events transmitted respectively. 
\begin{equation}
    ER = \frac{N_{s} - N_{tx}}{N_{s}}
    \label{eq:ER}
\end{equation}

The second metric that we consider is the compression ratio (CR). This ratio is a commonly used metric and is used in other event-stream compression work such as \cite{khan2020lossless}. CR is defined as the number of bytes in the original data stream divided by the size of the compressed stream in bytes. For this metric, we must consider the additional data required for each event that includes flow information. The equation for the calculation of CR is shown in \eqref{eq:CR}.

\begin{equation}
    CR = \frac{N_{s}\times8}{(N_{tx} - N_{nf})\times11 + N_{nf}\times8}
    \label{eq:CR}
\end{equation}

For \eqref{eq:CR} and the rest of this research, it is assumed that flow data is represented as two 12-bit values. This requires an additional three bytes to be transferred for each event containing flow. Therefore, flow events are 11 bytes in total while normal AER events are eight bytes as used in \cite{khan2020lossless,banerjee2021lossy}. The value of $N_{nf}$ equals the number of events in the event stream for which no flow can be calculated. All of these events must be transmitted, but they do not require extra bytes because they have no flow. The difference between the total number of events transmitted and the no-flow events is the number of events with flow that require an additional three bytes of information. Except where explicitly stated for cascaded compression, we assume that no additional compression is applied to the event stream beyond the FBC method presented.

\subsubsection{Event-Stream Similarity}
\label{sec:event-stream-distance}

Because the FBC method introduced is a form of lossy compression, we must evaluate the performance in terms of event-stream reconstruction accuracy. There are various methods presented in the literature that evaluate the similarity of two event streams. Some accumulate event frames and use standard image similarity metrics such as structural similarity index measure (SSIM) and peak signal-to-noise ratio (PSNR). These metrics are used in \cite{banerjee2021lossy}, however, they acknowledge that this method gives little insight into temporal errors in the reconstructed event stream. In fact, PSNR and SSIM have substantial limitations even for accumulated event frames. Both metrics are designed to evaluate traditional images where each pixel has an intensity value within some dynamic range. The task of quantifying event-stream similarity is substantially different. In this case, the data is sparse (i.e., not all pixels have data) and the events are binary without any explicit intensity information. Therefore, direct use of either SSIM or PSNR on event-stream data would have limited utility and produce potentially misleading reconstruction quality results.

The limitations of traditional image-based methods motivate the need for a metric specifically developed for quantifying the similarity between two event streams while taking into account the unique characteristics of asynchronous event data. This metric should capture changes both spatially and temporally and be applied to event streams of different lengths. For this research we use the asynchronous spatiotemporal spike metric (ASTSM) proposed in \cite{li2021asynchronous}. This method applies a spatiotemporal Gaussian to the event streams to convert them into a reproducing kernel Hilbert space (RKHS). By computing the inner product of the event streams in the RKHS the distance between the two streams can be found. This event-stream distance will be our metric of similarity used in evaluations. For a more detailed discussion on this method please refer to \cite{li2021asynchronous}. 

To implement this method we utilize the open-source code provided by the authors of \cite{li2021asynchronous}. The algorithm takes the parameters $\sigma_{x}$, $\sigma_{y}$ and $\sigma_{t}$ for constructing the Gaussian kernel. For computational traceability, the event streams are split spatially and temporally into `event cubes' between which the distances are calculated. These cubes have parameters $W$, $H$, and $L$ for width, height, and temporal length. For all metrics reported in this research we use ($\sigma_{x}=5 px$, $\sigma_{y}=5 px$, $\sigma_{t}=5000 \mu s$) as used in \cite{li2021asynchronous}. For the event cubes, we make the spatial size equal to that of the sensor and use 5 $ms$ as the value of $L$. For interpretability of the results, we normalize the distance computed using the ASTSM method for each pair of event cubes by the number of events in the original stream that fall into that cube.

\begin{figure*}[h]
    \centering
    \includegraphics[width=\linewidth]{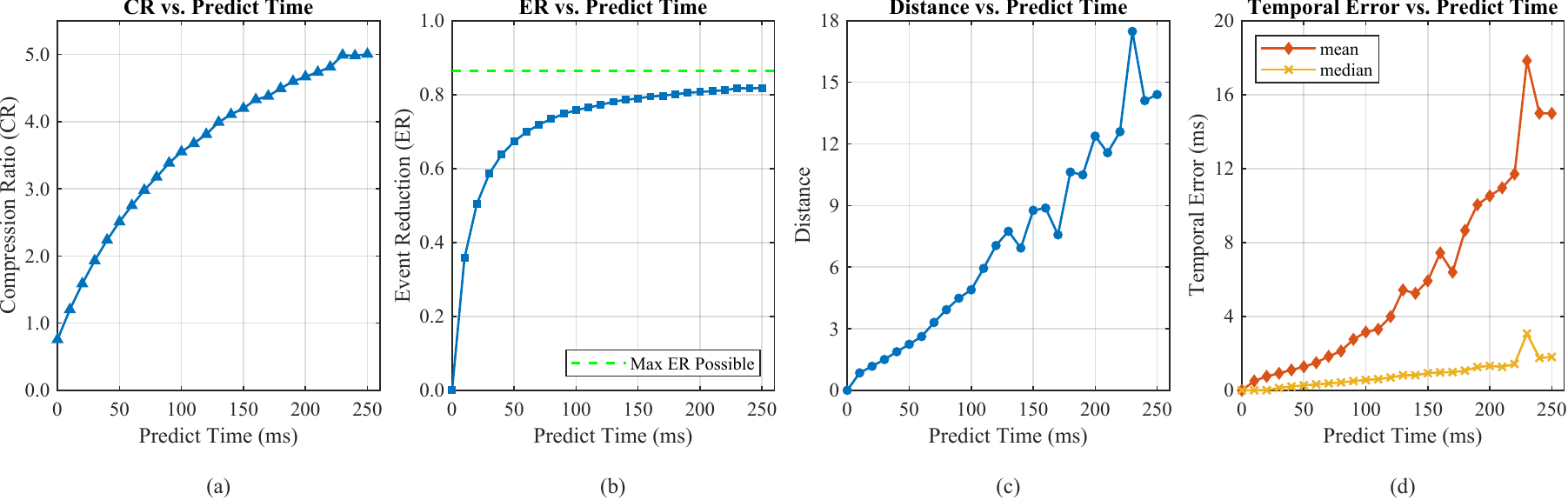}
    \caption{Characterization of FBC performance across various values of $PT$. The metrics evaluated are CR (a), ER (b), distance (c), and temporal error (d). Results use the bar-square dataset for performance characterization.}
    \label{fig:bs-characterization}
\end{figure*}

\subsubsection{Temporal Error}
We also evaluate results in terms of the temporal error, apart from the spatial error included in the distance metric. This temporal error is evaluated on a local pixel level by matching events at a given pixel to the temporally closest event within a 3$\times$3 window of that same pixel. Both the mean and median temporal error (TE) over a given dataset are reported. This metric is primarily used for the evaluation of the overall temporal accuracy of the method for a given scenario. While it does not explicitly include spatial information, it is important to note that spatial and temporal errors are often produced by the same source, namely errors in flow estimates.

\section{Experiments and Results}
\label{sec:experiments-and-results}

In the following sections, we evaluate the performance of the FBC method introduced. We first characterize how changing the timing parameters impacts the evaluation metrics established in Section~\ref{sec:evaluation-metrics}. An example of how the reconstruction performs over time when applied to a dataset is then shown. Finally, FBC is applied to a variety of real-world datasets and the results are presented and discussed. The default configuration of the algorithm pixel slack variable $\xi$ is empirically set to 0.4 for all results.

\subsection{Parameter Characterization}
\label{sec:parameter-characterization}

To characterize how changes in the $PT$ impact the evaluation metrics, the bar-square dataset used in \cite{Akolkar2020Real-timeFlow,stumpp2021harms} is used. This dataset includes a qVGA-resolution event stream with a bar and square oscillating vertically. This dataset was chosen for characterization due to the proven ability to compute accurate flow for the objects in the scene. Additionally, it is not as simple as some datasets that have one consistent direction of motion across the entire dataset. The fARMS algorithm with Savitzky-Golay plane-fitting local flow from \cite{Rueckauer2016EvaluationSensor} is used to compute the optical flow in order to achieve denser flow results. Of the 1.26 million events in the dataset, a valid flow estimate is found for 1.09 million of them. This means that at least 170 thousand events must be sent from this event stream, corresponding to a maximum possible ER of 0.86.

Fig.~\ref{fig:bs-characterization} shows the changes in CR, ER, distance, and temporal error as the value of $PT$ changes. Adaptive configuration of the send time as discussed in Section~\ref{sec:param-config} is used and results in values averaging $\sim8$ ms. This results in the number of events produced being within as little as 0.15\% of the original stream. For larger values of $PT$ where prediction performance decays, this difference can climb to as high as 13.5\%, however, it remains low for most reasonable operating points. As expected, increasing the prediction time increases CR and ER, but at the cost of increased event-stream distance and temporal error. This behavior is the predictable result of reduced data transmission and increased prediction durations. The CR begins below one due to the overhead of transmitting flow information in addition to event data. The break-even point where $CR = 1$ occurs at $PT = 5~ms$. For both CR and ER the best gains are achieved when initially increasing $PT$ and then marginal returns eventually begin to decline.

The distance metric shown in Fig.~\ref{fig:bs-characterization}(c) steadily increases until reaching a predict time of 130~ms. Beyond this point the upward trend in the event-stream distance continues, however, it is nosier and no longer monotonically increasing. This behavior is due to the nature of the bar-square dataset and the quality of flow during the $sending$ state. The dataset includes periodic oscillations of the objects in the scene. When the objects are changing direction the flow estimates are poor. For large values of $PT$ where extended predictions are being performed, the quality of flow has a significant impact on the resulting event-stream distance. Therefore, a higher $PT$ can result in a lower distance if the quality of the flow during the $sending$ states is higher. This behavior can be observed at any value of $PT$, however, the nature of the bar-square dataset only makes it apparent for larger values of $PT$.

The mean temporal error shown in Fig.~\ref{fig:bs-characterization}(d) follows similar trends as the distance metric. This trend is to be expected as the temporal error is a component of the total spatiotemporal distance between the reconstructed and original event stream. Comparison of the mean temporal error with the median temporal error shows that the median temporal error increases at a much slower, and more linear rate as $PT$ is increased. This behavior is due to the large effect that outliers have on the mean temporal error. Occasionally, when flow estimates have significant error, events will be predicted in areas of the frame where no ground-truth events will occur for a significant amount of time, thus leading to large temporal-error outliers. While both mean and median temporal error are informative of behavior, the median better estimates typical performance due to its robustness to outliers. Overall, the results in Fig.~\ref{fig:bs-characterization} demonstrate the wide range of compression ratios and event reduction that may be achieved. The exact configuration of the FBC algorithm should be determined based on the use case and required reconstruction accuracy.

\subsection{Temporal Behavior}
\label{sec:temporal-behavior}
\begin{figure}
    \centering
    \includegraphics[width=\linewidth]{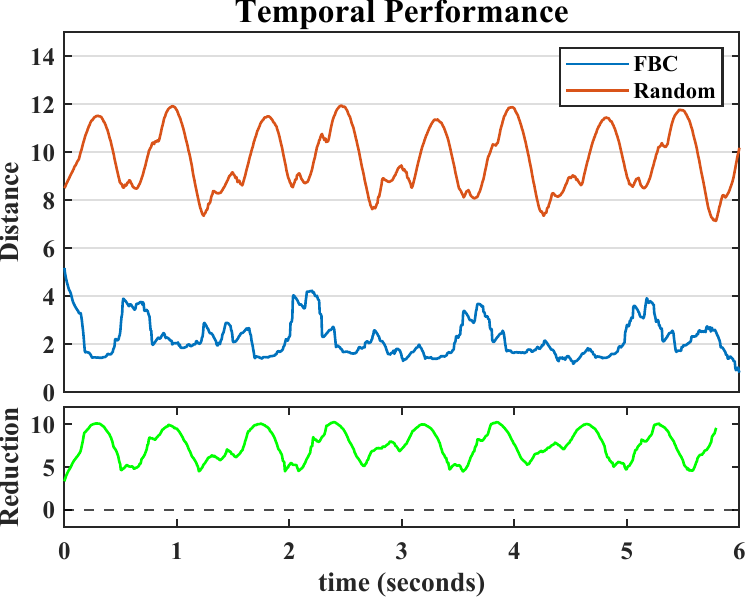}
    \caption{Reconstruction performance over time compared to equivalent ER achieved with random removal of events. Distance values are computed in $5~ms$ cubes and displayed as a $250~ms$ moving average of samples to reduce noise.}
    \label{fig:temporal-behavior}
\end{figure}

In addition to characterizing the overall performance of the FBC algorithm, it is also important to evaluate the behavior of FBC temporally as compression is being performed. Fig.~\ref{fig:temporal-behavior} shows the reconstruction performance over time compared to a randomly reduced event stream. For this evaluation, $PT$ is set to be $50 ms$. This configuration achieves an ER and CR of 0.67 and 2.51, respectively. The randomly reduced event stream is generated by randomly removing events from the stream, such that the ER equivalent to the FBC method (0.67) is achieved.

Fig.~\ref{fig:temporal-behavior} shows that the FBC distance is consistently less than the randomly reduced stream. Additionally, periodic behavior is observed due to the oscillation of the objects in the bar-square dataset. Local minima in FBC distance occur when the objects are changing direction. At these moments it is difficult for flow to be estimated, so most of the events are being sent and are therefore exact matches to the original stream. These periods where a low proportion of events get flow estimates could result in available bandwidth being saturated by the event rate. However, these moments are typically characterized by low overall event rates and are transitory by nature, meaning the overall system impact is temporary. The observation of these moments where most events are being sent validates the assumption that events without flow should be sent to enable proper event-stream reconstruction. On average, the FBC method has a 76.89\% lower distance over the duration of the stream while also preserving more event data in the reconstructed stream.

\subsection{Real-World Dataset Performance}
\label{sec:real-world-dataset-performance}

\begin{table}[b]
\setlength{\tabcolsep}{3pt}
\small
\centering
\caption{FBC performance on real-world datasets using $PT = 30~ms$.}
\begin{tabular}{@{}lccccc@{}} 
\toprule
\multicolumn{1}{c}{\textbf{Dataset}} & \begin{tabular}[c]{@{}c@{}}\textbf{CR}\end{tabular} & 
\begin{tabular}[c]{@{}c@{}}\textbf{ER}\end{tabular} &
\begin{tabular}[c]{@{}c@{}}\textbf{Distance}\end{tabular} &
\begin{tabular}[c]{@{}c@{}}\textbf{mean TE}\end{tabular} &
\begin{tabular}[c]{@{}c@{}}\textbf{med TE}\end{tabular} \\ 
\midrule
Shapes \cite{Mueggler2017TheSLAM}  & 3.62 & 0.79 & 6.18 & 2.27 & 0.37 \\ 
Bar-Square \cite{Akolkar2020Real-timeFlow} & 1.93 & 0.59 & 1.50 & 0.92 & 0.13\\ 
Outdoor Day 1 \cite{Zhu2018ThePerception}  & 2.89 & 0.71 & 3.76 & 3.25 & 0.64\\ 
Indoor Flying 1 \cite{Zhu2018ThePerception}  & 1.93 & 0.53 & 1.59 & 5.86 & 0.40 \\ 
Slider Far \cite{Mueggler2017TheSLAM} & 3.70 & 0.79 & 2.41 & 3.38 & 0.85 \\ 
\midrule
\textbf{Averages} & \textbf{2.81} & \textbf{0.68} & \textbf{3.07} & \textbf{3.14} & \textbf{0.48} \\
\bottomrule
\end{tabular}
\label{table:real-world-FBC}
\end{table}

\begin{figure*}
    \centering
    \includegraphics[width=.95\linewidth]{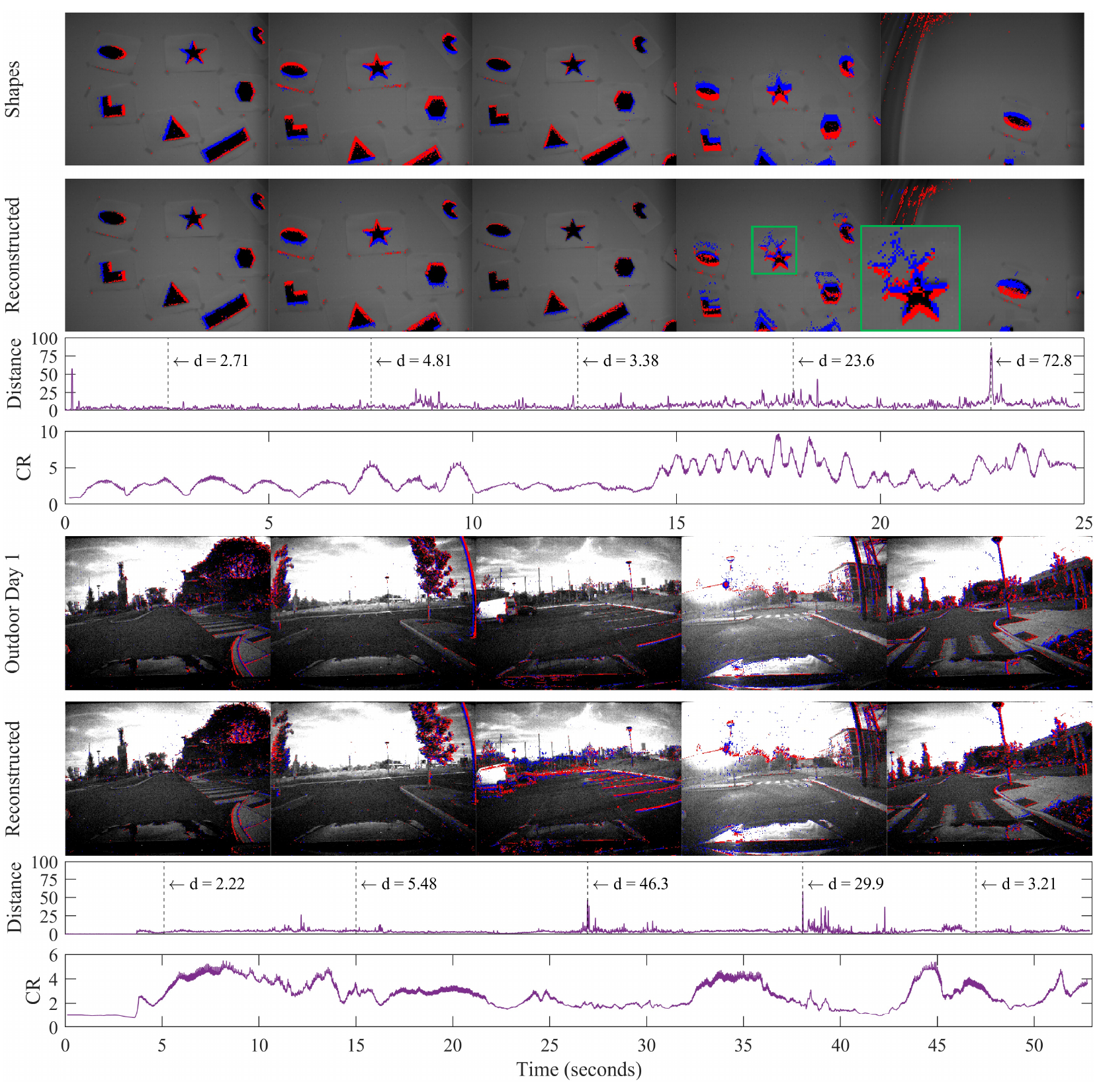}
    \caption{Qualitative FBC results for Shapes (top) and Outdoor Day 1 (bottom) scenarios. Events are accumulated for 10~$ms$ (top) and 16~$ms$ (bottom). ON events are in blue and OFF events are in red. Event stream distance at each sample time is annotated as $d$. In column 5 (green box) we see a larger qualitative error corresponding to a higher event distance. This error is the result of an instance in which the assumptions underlying the event prediction algorithm temporarily do not hold. The compression ratio is computed for a sliding 200~$ms$ window for each scenario.}
    \label{fig:qual-fbc-results}
\end{figure*}

Additional experiments were performed on various real-world datasets to evaluate the performance across different scene dynamics. The results of these experiments are shown in Table~\ref{table:real-world-FBC}. These results were collected using FBC configured with $PT = 30~ms$. The results show that across varied datasets the FBC method achieves an average CR of 2.81, which corresponds to the removal of 68\% of events on average. This method results in an average distance of 3.07 and a median temporal error of 0.48 ms. This temporal error represents only 1.6\% of the total prediction time. Optical flow parameters were lightly tuned based on the characteristics of the dataset. The Indoor Flying 1 dataset has the lowest compression ratio, which is caused by the fact that only 6.4 million of the 12 million total events in that dataset received an optical-flow estimate. A combination of factors such as the scene dynamics, flow computation parameters, and sensor noise causes a low percentage of events with flow. The Shapes dataset has by far the highest distance of all recorded results. This characteristic is due to specific portions of the overall scene which have extremely high event rates with rapid changes in sensor direction and orientation. This nature results in errant flow estimates and subsequent predictions. The Shapes and Slider Far datasets achieved comparable CRs. Slider Far, however, had a 61\% lower distance, but a 49\% higher mean temporal error. This characteristic indicates that errors introduced by FBC to the Shapes dataset were more spatial in nature compared to those introduced to the Slider Far dataset. It is also important to note that the motion of objects into and out of the sensor frame is another source of error throughout. It is impossible to predict events coming into the frame if they were outside of the frame during the $sending$ state. This error is magnified for faster and more dense data.

Fig.~\ref{fig:qual-fbc-results} shows qualitative results for the Shapes and Outdoor Day 1 scenes. Overall, reconstruction quality is quantitatively high. However, for frames with distances greater than 20, errors in reconstruction can be seen. For example, the fourth panel of the Shapes dataset shows clear artifacts when looking at the magnified star. In this instance, the error is caused by a rapid change in direction during the $predicting$ state. This is an instance where the assumption of constant motion used for the prediction algorithm fails to hold. In panel three of the Outdoor Day 1 scene, a high distance is also seen. In this case, the error is due to there being significantly more predicted events than actual events. This is likely caused by an overestimation of the required send time in the previous $sending$ state. Similar behavior is seen in panel four. Although there are too many events generated, they still accurately reflect the underlying scene. In both scenarios, the compression ratio fluctuates based on scene dynamics and the quality of the flow. The largest drops in CR occur when flow cannot be accurately calculated for a large portion of events generated, leading to high transmission requirements even in the $predicting$ state.

These results demonstrate that the FBC method can be successfully applied to a wide range of visual scenes and achieve compression performance comparable to existing state-of-the-art algorithms. For example, spike-coding methods achieved CRs of 3.84 and 3.78 for the Slider and Shapes datasets, respectively \cite{khan2020lossless}. This compression performance is only slightly above what is demonstrated for FBC using a relatively low $PT$ of 30$~ms$. Additionally, the stream-to-stream nature of FBC enables a cascaded compression pipeline to be realized for even further compression gains. When using this cascaded approach, state-of-the-art performance can be achieved as demonstrated in the next section.

\subsection{Cascaded Compression Performance}
\label{sec:combination-method}

In this section, we consider a scenario where real-time asynchronous compression is not required, and instead, the objective is to achieve maximum CR for reduced data size. In this case, the cascaded compression method with LZMA compression is utilized. The same datasets used in Section~\ref{sec:real-world-dataset-performance} are used here. The compression ratio achieved for each dataset when using FBC+LZMA is reported when $PT$ is 30, 60, and 90 milliseconds. As discussed in the FBC parameter characterization of Section~\ref{sec:parameter-characterization}, these predict times are reasonable and can provide low event-stream error depending on the quality of flow and dynamics of the scene. The results of this experiment are reported in Table~\ref{table:FBC+LZMA}. On average an additional 3.72$\times$ compression is achieved for CR-30 results. The maximum CR of 29.16 achieved is when $PT = 90$ for the Slider Far dataset. This dataset is particularly amenable to further compression as it has a roughly constant direction and velocity of motion, meaning there is redundant information in the flow that may be leveraged by LZMA. Overall, these results demonstrate the significant opportunity for further compression that is provided by the stream-to-stream nature of the FBC method.

\begin{table}[t]
\small
\centering
\caption{Achieved CR for cascaded compression architecture using LZMA. Results are reported for three different values of $PT$, denoted CR-$PT$. }
\begin{tabular}{@{}lccc@{}} 
\toprule
\multicolumn{1}{c}{\textbf{Dataset}} & 
\begin{tabular}[c]{@{}c@{}}\textbf{CR-30}\end{tabular} & 
\begin{tabular}[c]{@{}c@{}}\textbf{CR-60}\end{tabular} & 
\begin{tabular}[c]{@{}c@{}}\textbf{CR-90}\end{tabular} \\ 
\midrule
Shapes \cite{Mueggler2017TheSLAM}           & 13.69 & 18.71 & 23.91 \\ 
Bar-Square \cite{Akolkar2020Real-timeFlow}  & 8.11 & 11.18 & 13.43 \\ 
Outdoor Day 1 \cite{Zhu2018ThePerception}   & 9.26 & 11.90 & 13.16 \\ 
Indoor Flying 1 \cite{Zhu2018ThePerception} & 5.51 & 6.23 & 6.54 \\ 
Slider Far \cite{Mueggler2017TheSLAM}       & 15.69 & 23.46 & 29.16 \\ 
\midrule
\textbf{Averages} & \textbf{10.45} & \textbf{14.30} & \textbf{17.24} \\
\bottomrule
\end{tabular}
\label{table:FBC+LZMA}
\end{table}

\subsection{Receiver Run-Time Evaluation}

The final experiment performed was an evaluation of the receiver run-time. The receiver complexity comes from two main components: event prediction and event sorting. We perform a parameterized evaluation of both of these steps that make up the overall latency of the system for both a desktop-grade and an embedded platform. The receiver algorithm is written in C++ and compiled with ``-03" optimization. Desktop-grade performance results are collected on a 12th Generation Intel(R) Core(TM) i7-12700K CPU using one core. Embedded performance results are collected on a Quad-Core ARM Cortex A53 at 1.2 MHz using four cores with OpenMP parallelization. For performance evaluations, the events to predict are randomly generated with velocities ranging from -1000 to 1000 pixels/second and with 640$\times$480 (VGA) resolution. However, the resolution does not impact prediction performance. The system is considered to be achieving real-time performance if the receiver latency is less than the configured value of $PT$.

\begin{figure}[t]
    \centering
    \includegraphics[width=\linewidth]{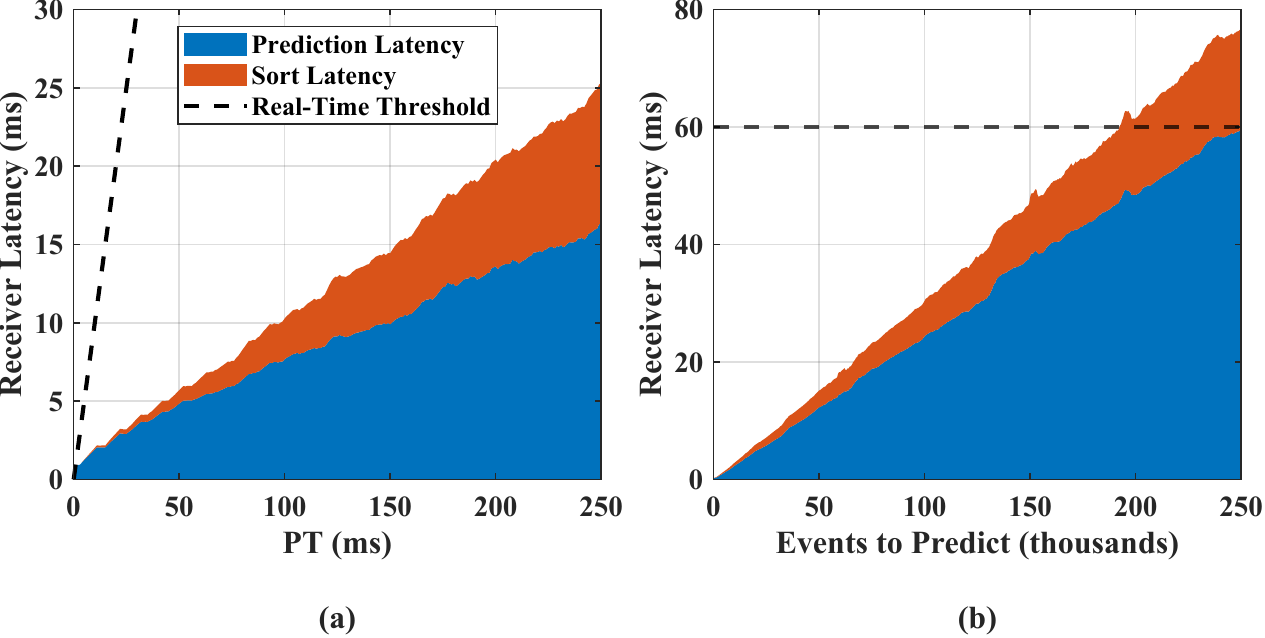}
    \caption{Desktop-grade receiver latency results for sweep of $PT$ (a) and sent events to predict (b). Figure (a) uses a fixed number of sent events to predict of 25,000 and (b) uses a fixed $PT$ of 60$~ms$.}
    \label{fig:latency}
\end{figure}

\begin{figure}[t]
    \centering
    \includegraphics[width=\linewidth]{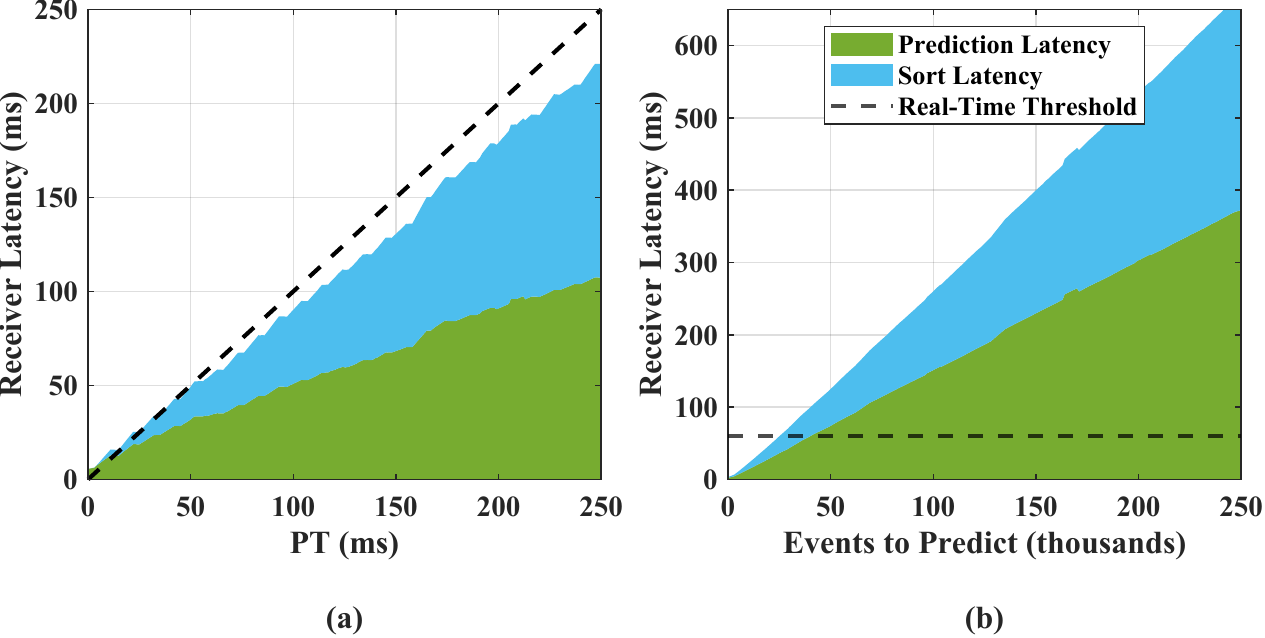}
    \caption{Embedded receiver latency results for sweep of $PT$ (a) and sent events to predict (b). Figure (a) uses a fixed number of sent events to predict of 25,000 and (b) uses a fixed $PT$ of 60$~ms$.}
    \label{fig:embedded-latency}
\end{figure}

Receiver latency is first characterized with respect to changing values of $PT$. Increasing $PT$ increases the number of candidate prediction points produced by the modified Bresenahm algorithm, therefore increasing prediction latency and total number of output events. This increased latency is shown in Fig.~\ref{fig:latency}(a) and Fig.~\ref{fig:embedded-latency}(a) where the number of sent events to predict forward is fixed at 25,000. It is observed that the actual achieved latency scales much slower than the real-time threshold for the desktop-grade device. Therefore, real-time performance is easily achieved even without algorithm parallelism. Real-time performance is enabled even with the embedded device, however, there is much less margin compared to the desktop-grade results. Fig.~\ref{fig:latency}(b) and Fig.~\ref{fig:embedded-latency}(b) show how the receiver latency scales for desktop- and embedded-grade devices respectively when the number of sent events to predict forward increases. In this case, the prediction time is fixed at 60$~ms$. The desktop-grade system maintains real-time performance until the number of events to predict forward hits 200,000. This threshold is an extremely large number of events that is more than an order of magnitude larger than the number of events observed for the datasets used in this research. This disparity indicates that the system has a significant margin to scale to higher-resolution sensors and high-activity scenarios while continuing to maintain real-time prediction and stream-reconstruction performance. The embedded system fails to achieve real-time performance beyond 25,000 events. This still enables real-time performance on the datasets evaluated in this research and demonstrates the capability for communication between edge devices. Further acceleration and improved embedded architectures will enable this embedded real-time threshold to be increased in the future. Although the theoretical system complexity is estimated to be $O(n^2\log(n))$, in practice we observe that the total complexity scales roughly linearly in the single-core design due to small values of $n$ and the dominance of prediction latency over the sort latency.

\section{Conclusion}
\label{sec:conclusion}

In this research, we introduced an event-stream compression method based on the use of asynchronous event-based optical-flow estimates to predict future events. This flow-based compression method leverages short periods of flow information to predict future events and reduce the total number of events and data that must be sent between transmitter and receiver, thus enabling reduced communication bandwidth and power requirements. The FBC method was characterized and evaluated on a variety of event-stream datasets and was found to achieve an average compression ratio of 2.81 while enabling a total event reduction of 68\% on average. This compression corresponded to a median temporal error of 0.48 $ms$ and a spatiotemporal distance of only 3.07 on average. Additionally, the unique stream-to-stream nature of the FBC method enabled further compression of up to 29.16$\times$ to be achieved through the use of LZMA dictionary encoding for lossless compression. The use of such a cascaded compression technique enabled state-of-the-art compression performance due to the compounding effects of leveraging multiple compression techniques. Furthermore, the latency of the receiver prediction algorithm was evaluated and demonstrated to be capable of performing predictions within real-time constraints for both embedded and desktop-grade platforms.

The results of this research demonstrate the benefits of using flow information for increased compression of event streams with minimal loss of information. The error that is introduced was measured by the use of the spatiotemporal event-stream distance metric and is caused by scenarios in which the fundamental assumptions of the algorithm outlined in Section~\ref{sec:event-prediction} do not hold. This error is most often caused by rapidly changing scene dynamics, errors in flow estimates, and sensor noise. In practice, due to the algorithm formulation, errors in the flow estimates have a particularly significant impact. Therefore, if improved optical flow algorithms are developed and used, then the benefits of FBC may continue to be leveraged, while further reducing the event-stream-reconstruction error. This adaptation promises to further enhance the already significant benefits that FBC can achieve for applications with communication bandwidth and power constraints through its enabling of accurate and fast compression of asynchronous visual event streams.

\bibliographystyle{./bibliography/IEEEtran}
\bibliography{./bibliography/ref}

\begin{IEEEbiography}[{\includegraphics[width=1in,height=1.25in,clip,keepaspectratio]{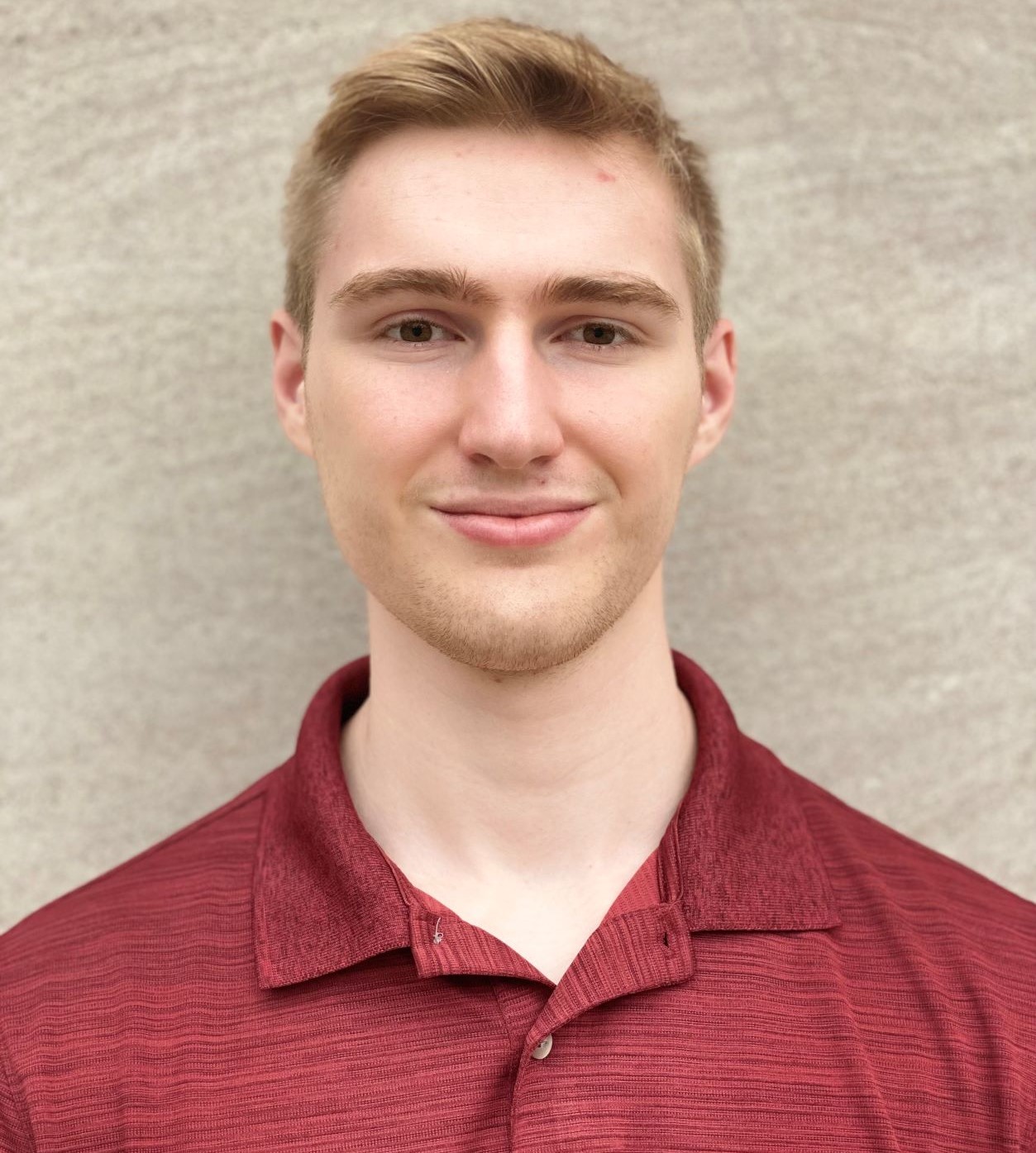}}]{\textbf{Daniel C. Stumpp}} 
received the M.S. degree in electrical and computer engineering in 2022 from the University of Pittsburgh where he is currently pursuing a Ph.D. degree in electrical and computer engineering. He is a member of the NSF Center for Space, High-Performance, and Resilient Computing (SHREC), where he performs research under the direction of Dr. Alan George. His research interests include high-performance FPGA-based accelerator architectures for embedded platforms, remote sensing applications of machine learning, and novel processing of asynchronous neuromorphic sensor data.
\end{IEEEbiography}

\begin{IEEEbiography}[{\includegraphics[width=1in,height=1.25in,clip,keepaspectratio]{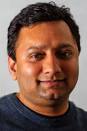}}]{\textbf{Himanshu Akolkar}}
received the M.Tech. degree in electrical engineering from IIT Kanpur, Kanpur, India, and the Ph.D. degree in robotics from IIT Genoa, Genoa, Italy, after which he had a postdoctoral position at Universite Pierre et Marie Curie. He is currently a Postdoctoral Associate with the University of Pittsburgh. His primary research interest includes the neural basis of sensory and motor control to develop an intelligent machine.
\end{IEEEbiography}

\begin{IEEEbiography}[{\includegraphics[width=1in,height=1.25in,clip,keepaspectratio]{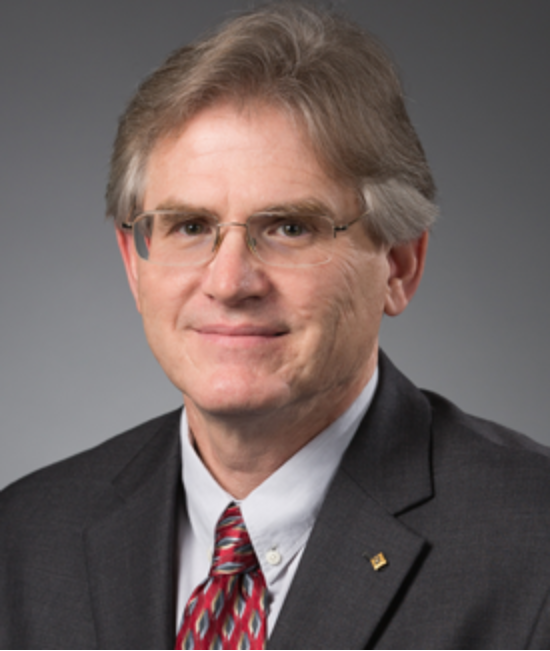}}]{\textbf{Alan D. George}}
(Fellow, IEEE) is currently the Department Chair, the Robert Horonjeff Mickle Endowed Chair, and a Professor of electrical and computer engineering (ECE) with the University of Pittsburgh. He is the Founder and the Director of the NSF Center for Space, High-Performance, and Resilient Computing (SHREC) headquartered at Pittsburgh. SHREC is an industry/university cooperative research center (I/UCRC) featuring some 30 academic, industry, and government partners and is considered by many as the leading research center in its field. His research interests include high-performance architectures, applications, networks, services, systems, and missions for reconfigurable, parallel, distributed, and dependable computing, from spacecraft to supercomputers. He is a Fellow of the IEEE for contributions in reconfigurable and high-performance computing.
\end{IEEEbiography}

\begin{IEEEbiography}[{\includegraphics[width=1in,height=1.25in,clip,keepaspectratio]{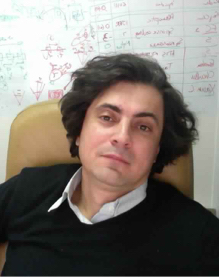}}]{\textbf{Ryad B. Benosman}}
received the M.Sc. and Ph.D. degrees in applied mathematics and robotics from University Pierre and Marie Curie, in 1994 and 1999, respectively. He is currently a Full Professor with the University of Pittsburgh/Carnegie Mellon/Sorbonne University. His work pioneered the field of event-based vision. He is also the Co-Founder of several neuromorphic related companies, including Prophesee—the world leader company in event-based cameras, Pixium Vision—a retina prosthetics company. He has authored more than 60 publications that are considered foundational to the field of event-based vision. He holds several patents in the area of event vision, robotics, and image sensing. In 2013, he was awarded with the national best French Scientific article by the publication LaRecherche for his work on neuromorphic retinas applied to retina prosthetics.
\end{IEEEbiography}

\end{document}